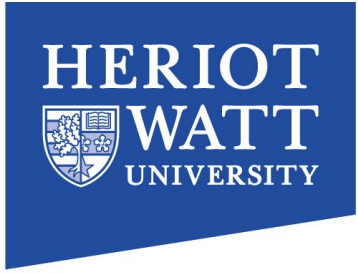

# RESEARCH ON THE CONCEPT OF LIQIUD STATE MACHINES

Name: Oladipupo Gideon Gbenga

email: ggo1@hw.ac.uk ; hgrv86@gmail.com

ABSTRACT

Liquid State Machine (LSM) is a neural model with real time computations which transforms the time varying inputs stream to a higher dimensional space. The concept of LSM is a novel field of research in biological inspired computation with most research effort on training the model as well as finding the optimum learning method. In this review, the performance of LSM model was investigated using two learning method, online learning and offline (batch) learning methods. The review revealed that optimal performance of LSM was recorded through online method as computational space and other complexities associated with batch learning is eliminated.

INTRODUCTION

A Liquid State Machines (LSM) is a computational model which consists essentially of recurrent and random spiking neural network and multiple read-out neurons (Zhang *et al*., 2015). Spiking Neural Network (SNN) according to Wall and Glackin (2013) is a third generation artificial neuron that is most biologically-inspired. They are preferred above the previous generations of artificial neurons because the spiking neurons operate in temporal domain and their computation is based on time resource. SNNs are becoming a dominant agent for brain-inspired neuromorphic computing - emulating the brain with computational hardware (Sharbati et al., 2018). The choice of SNNs is due to their inherent efficiency and accuracy on numerous cognitive tasks that includes speech recognition and image classification among others (Lee *et a*l., 2018).

According to Maass et al (2002), LSM model was developed from the viewpoint of computational neuroscience. An underlying principle of LSM lies in its ability of performing real time computations by transforming the time varying input stream into a higher dimensional space. LSM has three crucial components which are:
   (i) An input layer
   (ii) A reservoir or liquid and,
   (iii) A memoryless readout circuit.

The reservoir has numerous Leaky Integrate and Fire (LIF) neurons interconnected recurrently with biologically realistic parameters using dynamic synaptic connections. The readout is also implemented by several LIF neurons, however, they do not possess any interconnections within them. The liquid transforms the lower dimensional input stream to a higher dimensional internal state and these act as an input to the memoryless readout circuit which is responsible for producing the final output of the LSM. (Maass et al 2002). Figure 1 depicts a typical structure of LSM.

The concept of LSM was motivated by the versatility nature of neocortex (Yong and Peng, 2015). The neocortex is a key portion of the mammalian brain and through the networks of neurons, it controls functions such as sensory perception, motor command generation, spatial reasoning and conscious thought amongst others. The functionality of neocortex is based on the formation of six interconnected layers of neurons that are generated sequentially over a long time period (Christiana et al., 2016).

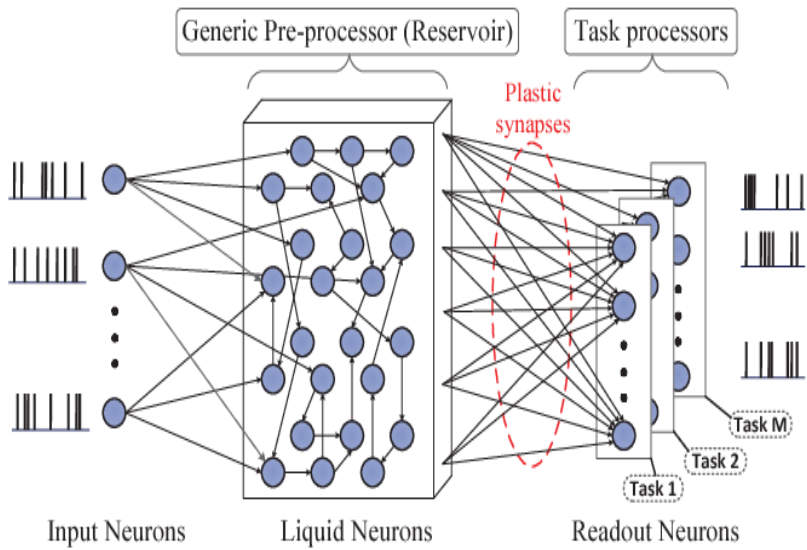

Fig. 1: A liquid state machine supporting multiple tasks.    (Wang *et.al* , 2015).

The focus of this research is to examine the significance of batch (offline) learning vis-à-vis online learning on LSM. In this regard, LSM model would be explored using three different papers and it would cover the description of the workings of LSM, its employment and mode of training and the results of batch and online learning method on LSM.

IMPLEMENTATION OF LIQUID STATE MACHINE

Jin and Li (2017), aligned with the three crucial parts of LSM which includes the input layer, the reservoir and the readout. The inputs signals of this LSM model was based on speech recognition. The signals were generated through speech signals preprocessed by a Lyon passive ear model (Jin and Li, 2017). The generated signals were subsequently passed through the Bens Spiker Algorithm (BSA), an algorithm for converting analog values to spike trains (Schrauwen and Campenhout, 2003). BSA was introduced to generate the spike trains as shown in Figure 2.

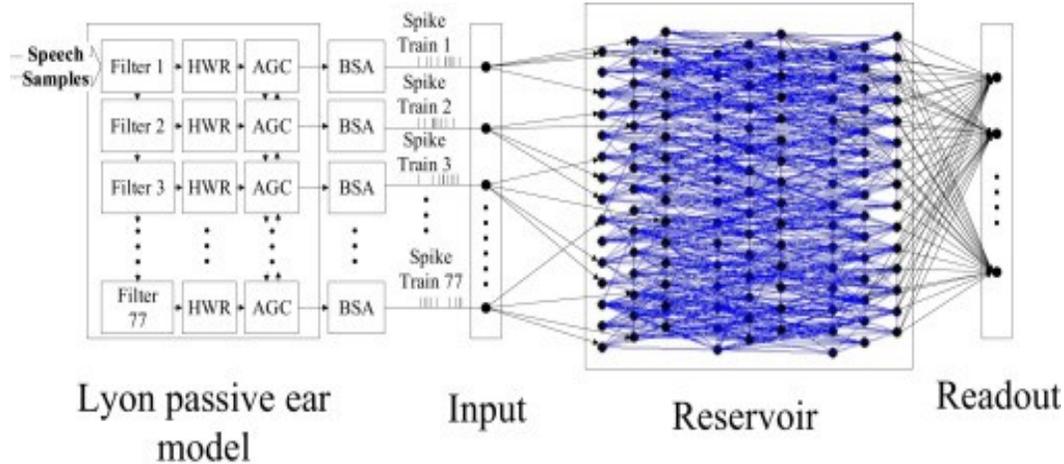

Fig.2. The LSM-based speech recognition system. (Jin and Li, 2017)

It identified two steps that were fundamental for the processing of input signals. In the first instance, the signal, which is the spike train *u(t)*, were mixed with responses from the LSM where the signals are transformed into a higher dimensional state. The reservoir via plastic synapse was connected to the readout in the second phase. The readout neuron at any time t receives from the reservoir a net current given by:

$$I_o(t) = \sum w_{oi} \cdot f_i(t) = \sum w_{oi} \cdot f_i[u(t)], \qquad (1)$$

where $f_i(t)$ is the response of the $i^{th}$ neuron in the reservoir, and $w_{oi}$ is the synaptic weight between the $i^{th}$ reservoir neuron and the readout neuron. The integrated net current over [0,T] is:

$$\int_0^T Io(t) \, = \sum w_{oi} \int_0^T fi(t) \, = \sum w_{oi} \int_0^T fi[u(t)]. \qquad (2)$$

The implication of the above is that the readout neuron will be taken as linear classifier of the reservoir responses since the integrated net current from the readout neurons is a linear combination of integrated outputs from reservoir neurons (Jin and Li, 2017).

De Azambuja *et al.* (2017) highlighted that the design of LSM was based on Short Term Plasticity (STP) which essentially has to do with dynamic changes in the synaptic efficiency over a period of time. STP refers to fast and reversible changes of synaptic strength caused by pre-synaptic spiking activity and it occurs on timescales from milliseconds to few seconds (Ghanbari *et al.*, 2017). The presynaptic spikes result in either facilitation where synaptic strength increases, or depression where synaptic strength decreases. Facilitation and depression are mediated by the dynamics of presynaptic calcium, depletion or the replenishment of vesicles in the presynaptic terminals (ibid).

De Azambuja *et al.* (2017) adopted an abstracted layer of LIF neurons as input and these directly injects into the reservoir. The virtual input layer was composed of 300 neurons subdivided into 6 groups. In this regard, Gaussian distribution which basically acts as modulator connects the virtual input layer and the reservoir. It modulates the weights between the pre-synaptic and post-synaptic neurons default standard deviation value, hence each input connection was spread creating a redundancy. The reservoir was created using the 600 LIF neurons (forming a 3 D structure of size 20 x 5) where 80% were excitatory and 20% inhibitory as shown in Figure 3.

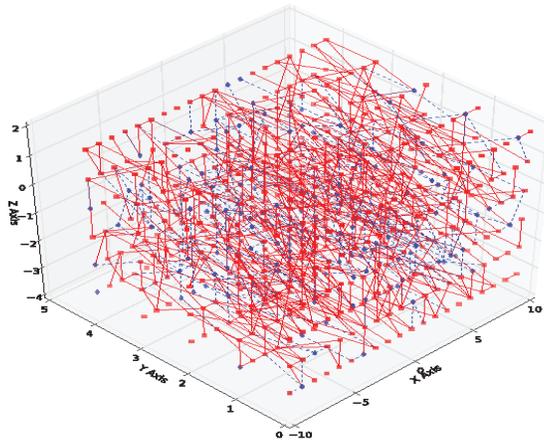

The readout neurons collect the generated high dimension space from the reservoir and interpret in linear form using a linear classifier. de Azambuja *et al.* (2017) LSM model is a biomimetic robot arm controller with proprioceptive feedback connections from the read out to the inputs. Figure 4 depicts the general architecture of the arm controller.

Fig.3. Visualisation of the liquid's shape. Red squares indicate the excitatory neurons while blue diamonds shows the inhibitory ones. Continuous red lines are excitatory connections and dashed blue lines inhibitory ones (de Azambuja *et al.*, 2017).

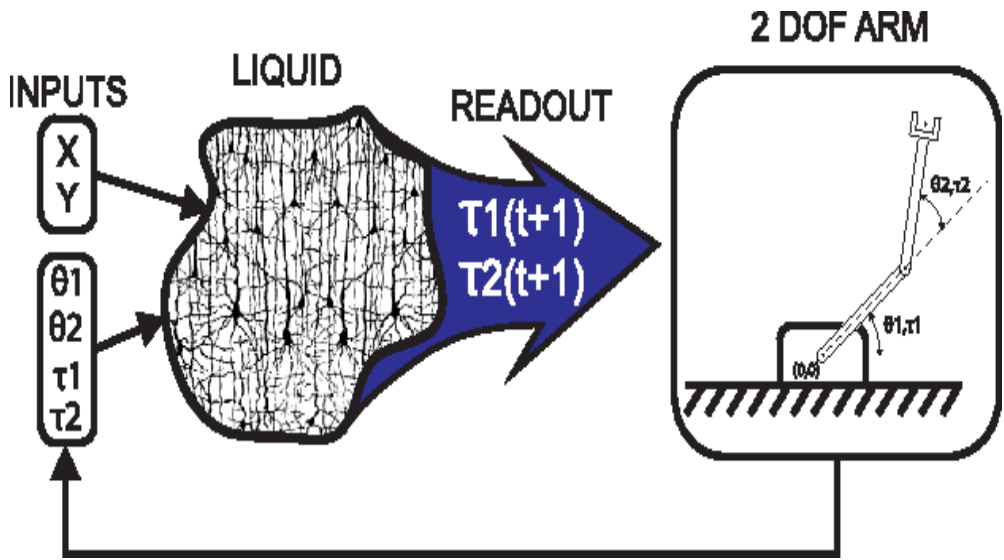

Fig.4. illustrative representation of the arm controller. X, Y are the coordinates of the final positions, T1 and T2 are the commanded torques and ө1 and ө2 the current angles (proprioceptive feedback) (de Azambuja *et al.,* 2017).

Avesani *et.al*, (2015) model of LSM was aimed at addressing the Hemodynamics Response Function (HRF) decoding task without incurring problems associated with HRF models such as Multivariate Pattern Analysis (MVPA) and Support Vector Machine (SVM) amongst others. In this model, the input signal *x(t)* consists of stimuli sequence, which essentially is called the Blood Oxygenation Level Dependent (BOLD) signal.

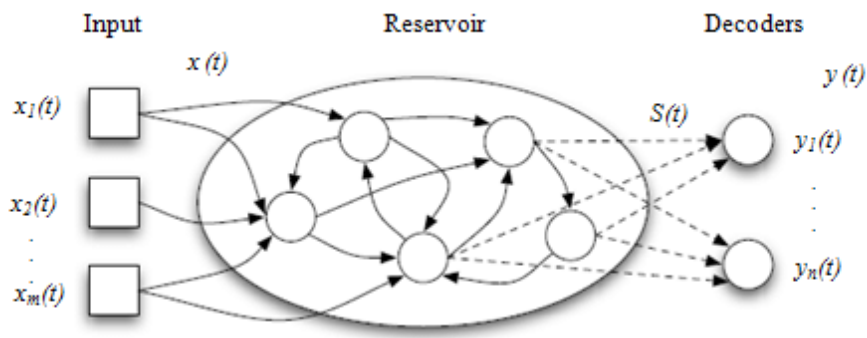

Fig.5 Reservoir computing network. The reservoir processes a multi-dimensional input data stream x(t) generating a series of high-dimensional internal state S(t). At the same time the decoders produce the required multi-dimensional output function y(t) based on the generated internal states (Avesani *et.al* 2015).

The reservoir according to Avesani *et.al*, (2015) is a non–linear system with a network of artificial neurons computed recurrently. Basically, it is used to encode the non-linear transformation of the input stream by capturing the past events and translating into a high-dimensional reverberating internal activity state. Conversely, the readout (decoder) retrieved the temporal information from reservoir and transforms it to a time-series output. The fundamental recurrent neural networks computation driving the reservoir is expressed as:

$$S(t) = L((x), S(t-1))$$
$$Y(t) = D(S(t)) \qquad (3)$$

where the internal network state $S$ at time $t$ is generated by an operator $L$ integrating the input value of $x$ to the network at current time $t$ with the previous internal state at time $t-1$. $D$ is the detector function which is commonly implemented using a clarification or a regression algorithm trained with simple learning mechanisms.

EMPLOYMENT AND MODE OF TRAINING THE LSM

The Jin and Li (2017) LSM's model focused its performance and robustness on speech recognition. It adopted a learning rule based on Hebbian principle in which the aim of the learning process was to regulate read out neurons in line with a stated desired level as well as the subsequent adjustments of the synaptic weights. In this model, the training set up was patterned after Zhang et al., (2015). In this case, each input spike train generated in the preprocessing stage is sent to four randomly chosen reservoir neurons through synapses with fixed weights. The weights are randomly chosen in the digital range of $W_{max}$ or $W_{min}$, where $W_{max}$ and $W_{min}$ are maximum and minimum synaptic weights used in the simulation, respectively. Plastic synapses connect reservoir neurons fully to each readout neuron and their weights are randomly initialized between $W_{max}$ and $W_{min}$. The plastic synapses are trained by the adopted learning algorithm (Jin and Li 2017).

The recognition performance was simulated using 5 randomly generated LSMs. The LSM was trained and tested according to the samples for five times with different training and testing datasets. For the i*th* ($i$ = 1, 2, 3, 4, 5) time, the i*th* group was used for testing and the remaining data for training. The recognition decision was made after each testing speech sample was played. At this time, the readout neuron that has fired most frequently is the winner and its associated class label is deemed to be the classification decision of the LSM. The five classification rates obtained in the testing stage were averaged as the final performance measure of the LSM (Jin and Li 2017).

The performance of LSM model of de Azambuja *et al.,* (2017) could only be verified when the task of reproducing four distinct trajectories and a biomimetic robot arm controller were implemented. The training considered four sets of experiments (Set A, B, C and D) and these were investigated using the four trajectories. A total of 320 simulations were carried out (20x4x4) for the training of the readouts. Also, the trials for each trajectory were executed by varying the noise levels and use of the STP. STP contributes to information transfer instead of frequency independent broadband behavior, hence, the need for noise to be filtered-out when STP is active. The default values for $i_{noise}$ and $i_{offset}$ as outlined in Table 1 were varied between sets A, B and C, D. Also the input variables receiving an additive noise were varied between sets A, B and C, D since the system receives feedback from the torques and joint angles. The trained readouts were tested through a total of 50 new trials for each one of the four trajectories. The results were used as the base to verify the impact of having STP enabled or not inside the liquid (de Azambuja *et al.*2017).

Table 1: Liquid Default Parameters (de Azambuja *et al.*2017)

| Parameter | Value | Unit |
|---|---|---|
| Membrane time constant ($\tau_m$) | 30.0 | $ms$ |
| Membrane capacitance ($C_m$) | 30.0 | $nF$ |
| Synapse time constant (exc. - $\tau_{syn_e}$) | 3.0 | $ms$ |
| Synapse time constant (inh. - $\tau_{syn_i}$) | 6.0 | $ms$ |
| Refractory period (exc.) | 3.0 | $ms$ |
| Refractory period (inh.) | 2.0 | $ms$ |
| Membrane Threshold | 15.0 | $mV$ |
| Membrane Reset | [13.8, 14.5] | $mV$ |
| Membrane Initial | [13.5, 14.9] | $mV$ |
| $i_{offset}$ | [13.5, 14.5] | $nA$ |
| $i_{noise}$ ($\mu$) | 0.0 | $nA$ |
| $i_{noise}$ ($\sigma$) | 1.0 | $nA$ |
| Transmission delay (exc.) | 1.5 | $ms$ |
| Transmission delay (inh.) | 0.8 | $ms$ |

Avesani *et.al*, (2015) LSM model focused on addressing the Hemodynamics Response Function (HRF) decoding task. The reconstruction of the Blood Oxygenation Level Dependent (BOLD) signals were trained and tested. The purpose of the testing phase was to evaluate the quality of the created model and thus to confirm/disprove the existence of any relationship between the stimuli and the recorded BOLD signal. In the training phase, a portion of data was given to a supervised learning process that fitted the readout/decoders' parameters to produce the required BOLD signal voxel by voxel given the sequence of stimuli. During the testing phase, a portion of the data was used as a hold-out set to generate the expected BOLD response related to a set of stimuli. This was then compared to the real BOLD signal for relevance analysis.

The original and the synthetic BOLD signals produced by the LSM were analyzed using two parameters which are the Root Mean Square Deviation (RMSD) and the Pearson correlation. When the RMSD value is low and the correlation value is high, it implies good prediction accuracy associated with high model quality and indicates the relevance of a voxel for a given cognitive task. Conversely, high RMSD and low correlation values implies low prediction accuracy, and this points to the fact that the model does not match the data (Avesani *et al*, 2015). With the adopted metric, it was easy to segregate relevant voxels from irrelevant ones.

RESULTS AND DISCUSSION

The performance of the LSM model of Jin and Li (2017) was tested with two additional subsets of the TI46 speech corpus, spoken words data set for training and evaluating speech recognition algorithms. Figure 5 analyzes the recognition rates of the LSMs on the adopted three different benchmarks. The complexity level 3 is the original setting adopted from Zhang *et al*., (2015) while complexity level 2 was the suggested setting of this model. Figure 6 shows clearly that the recognition rate degrades as the design complexity decreases for the three adopted benchmarks. The implication of this is that the optimum performance of the LSM in speech recognition by default depends solely on online learning as against batch learning since batch learning tends to reduce the complexity by providing prior information.

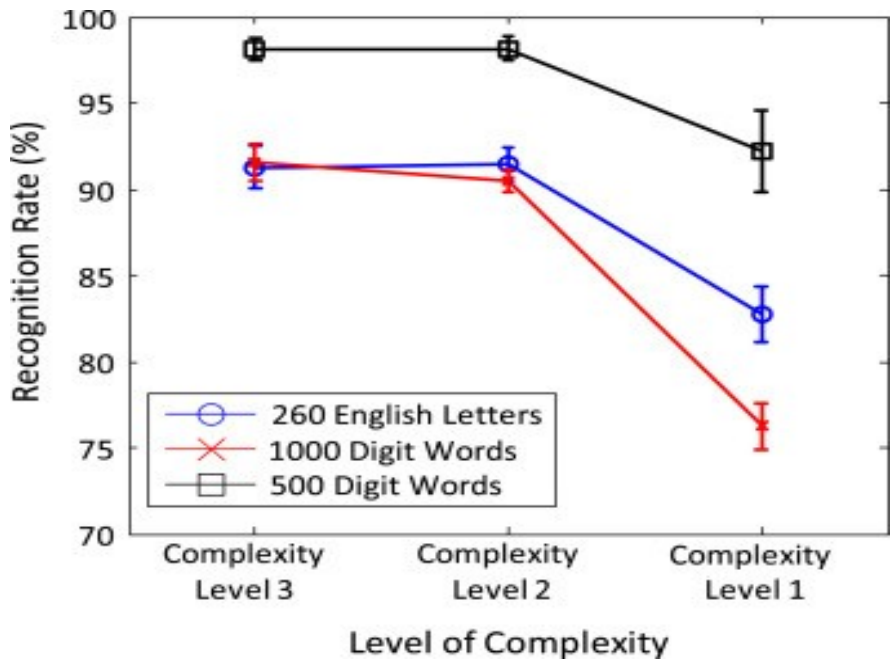

Fig.6 Classification performance of the LSM on the three adopted benchmarks decreases as a function of design complexity. (Jin and Li 2017)

The results from the Welch's t-test conducted on the model of de Azambuja *et al.* (2017) as shown in Figure 7 revealed that Sets A and C with STP performed better only during the trajectory 2 (triangles). For the trajectory 3, the results for all the sets were close to each other. There was no much difference for Sets A and B in trajectory 4.  However, Set D performed better than C here too. The results for trajectory 1 were always better when no STP was employed.  This paper equally affirmed that the LSM model performed better through online learning than through batch learning (de Azambuja *et al.*2017).

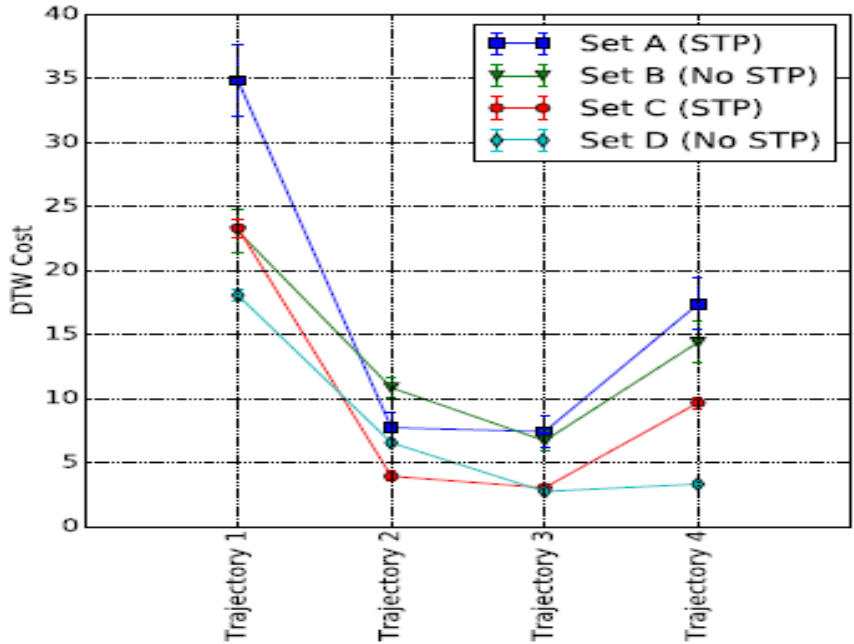

Fig.7 Dynamic Time Warping (DTW) cost considering the trajectories individually. The lower the DTW cost, the better are the results (de Azambuja *et.al* 2015).

The performance of LSM model as proposed by Avesani *et.al*, (2015) was analysed by correlating the behavior of the model based on predicted and observed pattern.  The results revealed high performance for both real data and synthetic BOLD signal where the voxels were 100% and 97% correctly determined respectively. This suggests that the brain maps could be generated with data-driven approaches, without requiring any prior knowledge on the expected HRF otherwise known as batch/offline learning.

One crucial aspect as noticed in the training and testing of these models is that LSM optimum functionality is achieved through online learning instead of batch/offline learning where all the data are available.  In offline learning, data are processed sample by sample and the parameters are updated on each new instance.  Also, offline learning facilitates a faster learning process and it is suitable when data changes with the time or when there is no access to all the training data in advance.  It is however worthy to note the online learning can cause residual errors when dealing with an outlier sample (Vallejo, 2018).

CONCLUSION

LSM is a neural model with real time computations which transforms the time varying inputs stream to a higher dimensional space. LSM has three crucial components which are an input layer, a reservoir or liquid and a memoryless readout circuit.  In this research views as outlined in the LSM models investigated reveals that its optimum performance is achieved when it is operated in its real and original states as against the suggested state.  In this regard, optimum performance is achieved through online learning in which there was no previous information to the LSM model on the specific tasks to be carried out. This in turn would reduce the implementation cost as well as unnecessary complexities that could be introduced through batch learning.   The insights gained would be helpful in practice as they offer a better and cheaper option for the implementation of robust LSM devoid of complexities.